# Anatomy-aware Self-supervised Learning for Anomaly Detection in Chest Radiographs

Junya Sato, Yuki Suzuki, Tomohiro Wataya, Daiki Nishigaki, Kosuke Kita, Kazuki Yamagata, Noriyuki Tomiyama, and Shoji Kido

*Abstract*—Large numbers of labeled medical images are essential for the accurate detection of anomalies, but manual annotation is labor-intensive and time-consuming. Self-supervised learning (SSL) is a training method to learn data-specific features without manual annotation. Several SSL-based models have been employed in medical image anomaly detection. These SSL methods effectively learn representations in several field-specific images, such as natural and industrial product images. However, owing to the requirement of medical expertise, typical SSL-based models are inefficient in medical image anomaly detection. We present an SSL-based model that enables anatomical structure-based unsupervised anomaly detection (UAD). The model employs the anatomy-aware pasting (AnatPaste) augmentation tool. AnatPaste employs a threshold-based lung segmentation pretext task to create anomalies in normal chest radiographs, which are used for model pretraining. These anomalies are similar to real anomalies and help the model recognize them. We evaluate our model on three opensource chest radiograph datasets. Our model exhibit area under curves (AUC) of 92.1%, 78.7%, and 81.9%, which are the highest among existing UAD models. This is the first SSL model to employ anatomical information as a pretext task. AnatPaste can be applied in various deep learning models and downstream tasks. It can be employed for other modalities by fixing appropriate segmentation. Our code is publicly available at: https://github.com/jun-sato/AnatPaste.

*Index Terms*—Self-supervised learning, unsupervised anomaly detection, chest radiograph, convolutional neural network, computer-aided diagnosis

## I. INTRODUCTION

DEEP learning has made significant progress in image recognition, including medical image recognition. A deep learning model learns informative image representations without manual feature extraction and preprocessing. It requires training with a large number of labeled images for accurate recognition [1]–[3]. However, manual annotation is labor-intensive and time-consuming. Moreover, medical image annotation requires medical expert intervention and resources [4]–[6].

Chest radiograph is the most frequently performed radiological examination in the world, with a reported 146 erect-view images acquired per 1,000 population annually [7]. They are used to detect and classify various cardiac and pulmonary disorders. Numerous images containing disease-specific annotations are needed for disease classification. In addition, identifying diseases whose images are not used during model training is challenging. Likewise, prediction accuracy becomes low when there is little data available, as in the case of rare diseases [8]. Hence, our goal is to solve the label deficiency and create a model with high diagnostic accuracy in chest radiographs.

An unsupervised anomaly detection (UAD) model is effective in such annotation and dataset size problems. This model learns only normal images and determines a characteristic threshold. The image exhibiting features beyond the threshold is classified as abnormal. We can easily collect normal images through medical examinations. Furthermore, UAD model can detect rare diseases unseen during model training. For these reasons, extensive studies have been conducted on UAD [9]–[19]. Recently, models using deep learning for feature extraction and models using encoder-decoder networks such as variational autoencoder (VAE) [12], [13] and generative adversarial network (GAN) [14]–[19] have been actively employed.

Self-supervised learning (SSL)-based UAD enhances anomaly detection accuracy [10], [19]–[22]. SSL aims at learning data-specific semantic features by generating supervisory signals from a pool of unlabeled data without human annotation [23]. In natural image classification datasets such as ImageNet [1], some SSL-based models have achieved accuracies comparable to conventional supervised learning [24]–[26]. SSL-based models require an appropriate pretext task to acquire detailed characteristics of diseases (anomalies) [27]. Several SSL-based deep learning models have been proposed in the medical field [19], [22], [23], [28]–[32]. However, these models use the same SSL as natural images and disregard task-specific information in medical images.

For example, some popular SSL-based methods generate abnormal images by random pasting [10], [21] or masking [19], [23], [26], [33]. Classifying or repairing them helps the model learn image features. The generated abnormal areas have clear boundaries with the normal ones and occur where they should not appear anatomically. These augmentations are inadequate for models to learn the characteristics of actual anomalies. Several lung diseases, such as lung cancer and pneumonia are

This study was supported by JSPS KAKENHI (grant numbers: JP21H03840).

Junya Sato, Yuki Suzuki, Tomohiro Wataya, Daiki Nishigaki, Kosuke Kita, Kazuki Yamagata, Noriyuki Tomiyama, and Shoji Kido are with the department of Radiology, Osaka University Graduate School of Medicine, 2-2 Yamadaoka, Suita-city, Osaka 565-0871, Japan (e-mail: {j-sato, kido} @radiol.med.osaka-u.ac.jp). Corresponding author: Shoji Kido.

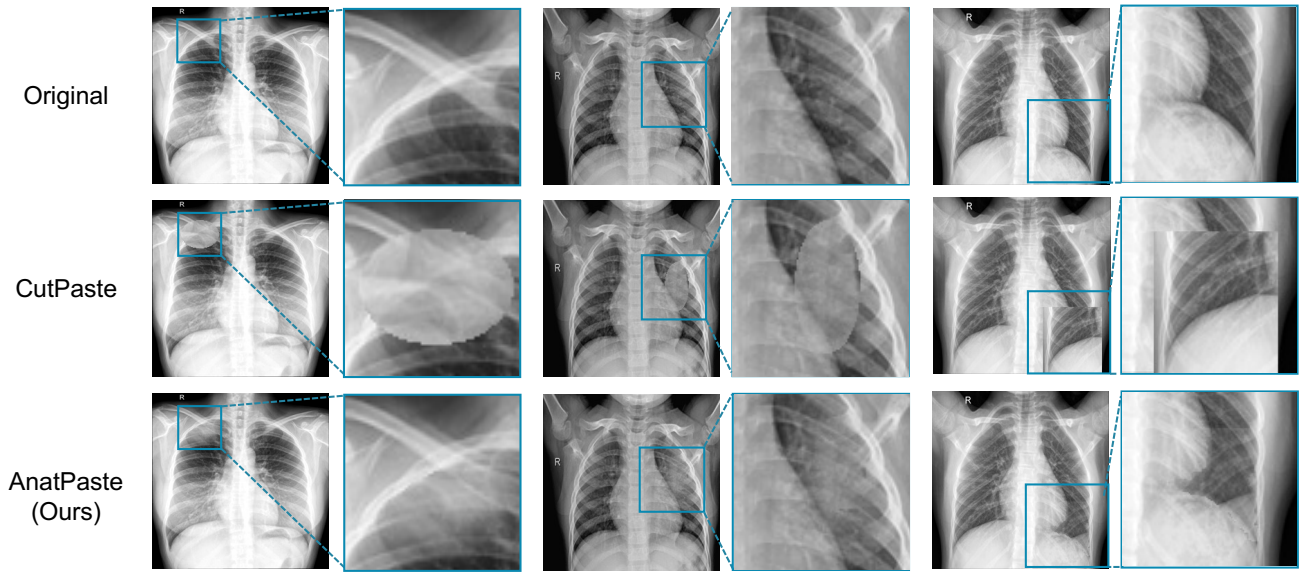

Fig. 1. Examples of the original chest radiographs (top row) that were augmented with either CutPaste [21] (second row) or AnatPaste (bottom row) algorithms. Zoomed in versions of the blue squared insets from each image are presented adjacent to the source images. The insets are in identical positions in each column.

detected using chest radiographs. These diseases result in anomalous shadows in the lung region and rarely in the abdomen or background (outside the body). Therefore, setting the SSL-based model to learn characteristic features from medical images is required for accurate anomaly detection.

We propose a novel SSL-based method, anatomy-aware pasting (AnatPaste) augmentation, which includes lung segmentation to exploit anatomical information. AnatPaste employs normal images and generates real anomaly-like images. This feature assists deep learning models to recognize anomalies. Fig. 1 depicts a comparison between AnatPaste and the existing SSL augmentation method. The existing method augments the images where anomalous areas are clearly demarcated and includes additional anomalous areas where the real anomalies do not occur. However, AnatPaste-developed anomalies appear real and are limited to the lung region.

AnatPaste does not require a specific deep learning model's structure; therefore, it will help existing UAD models boost their accuracy. In addition, if proper segmentation is performed, AnatPaste can be extended to other imaging modalities such as CT and MRI and organs. AnatPaste has the potential to be widely applied to medical image analysis.

To evaluate our method for anomaly detection, we use a simple one-class classifier trained with AnatPaste. The UAD model is validated using three publicly available chest radiograph datasets. It exhibits significantly higher anomaly detection accuracy on all the datasets compared with existing UAD models. The contributions of our study are summarized as follows:

- AnatPaste is the first augmentation method that utilizes medical image segmentation as a pretext task.
- AnatPaste can be applied to any model, downstream tasks, and even to other imaging modalities and organs by setting up segmentation methods.
- The model performs the best on anomaly detection tasks on three chest radiograph datasets.

## II. RELATED WORKS

### A. Anomaly Detection

Anomaly detection, also known as outlier detection or novelty detection, involves the detection of data instances that deviate significantly from majority instances [34]. Traditional models, such as One-Class SVM [35] and SVDD [35], train a single normal data class and determine a threshold to detect the outliers. However, these models increase computational complexity when used with large and high-dimensional data. Recent studies have increasingly employed deep learning models for feature extractions. VGG trained on ImageNet [1] is employed to obtain representations of X-ray security-screening images [36]. Deep SVDD [37] trains a representation to minimize the volume of a hypersphere of the normal samples.

Several models using encoder-decoder networks such as autoencoder have also been proposed to learn normal data distributions. AnoGAN [16] is the first generative adversarial network model for anomaly detection. AnoGAN identifies a query image as abnormal when there is a large difference between the query image and the image generated by learning only normal images. fAnoGAN [15] is a faster and more accurate version of AnoGAN owing to the introduction of an encoder. Ganomaly [38] and SALAD [19] improve anomaly detection accuracy using a loss function with reconstructed images and latent spatial variables. These models are effective on some datasets, but reconstruction errors make it challenging to capture minor anomalies [39], [40]. SQUID [18] employs space-aware memory queues, as medical images are structured owing to consistent imaging protocols. However, SQUID requires simultaneous training of its components (one discriminator, encoder, and two generators). Therefore, it encounters structural complexity.

### B. Self-supervised Learning

SSL is an approach that learns semantically useful features for a task through a pretext task without manual annotations. SSL generates a supervisory signal from a pool of unlabeled data for the downstream task. The key challenge for SSL is setting up an appropriate task that allows the model to learn the features of the dataset well. For natural images, various SSL methods have been proposed, such as contrastive learning [24], [25], rotation prediction [41], jigsaw puzzle [42], denoising autoencoder [43], and in-painting [33]. Although these methods have been widely used in medical imaging [19], [22], [23], [28]–[32], they just adopt the SSL method proposed for natural images without any modification. As an SSL method specialized for medical images, Zhang et al. [44] treat 3D images such as CT and MRI as a set of 2D slice images and perform a pretext task of predicting the spatial position of the 2D slices. Tian et al. [10] devise MedMix, a pretext task based on contrastive learning designed for medical images. However, they do not consider the anatomical structure of organs. CutPaste [21] is an SSL method specialized for anomaly detection. It generates an anomaly image by cutting a part of a normal image and pasting it to another location in the same image. It has achieved state-of-the-art performances in anomaly detection for the industrial product dataset MVTecAD [45]. However, many medical images have complex structures and often need medical expertise, making it challenging to determine anomalies. To achieve accurate anomaly detection in medical images, the method should be redesigned such that the model can understand the structure of complex anatomy.

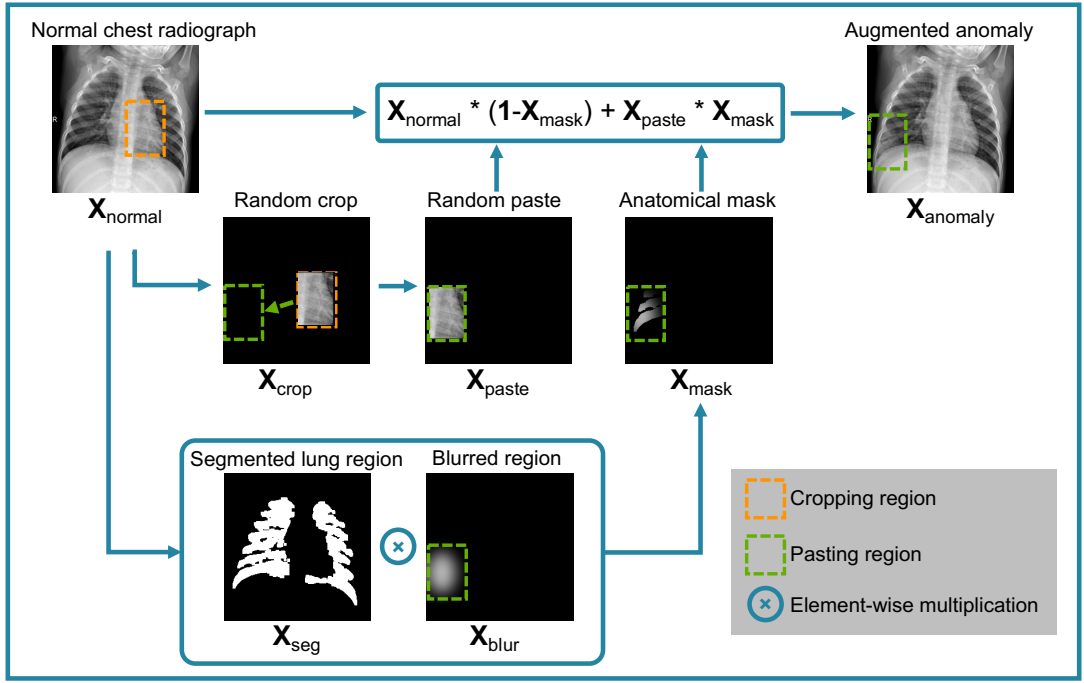

Fig. 2. Illustration of AnatPaste augmentation. A patch ($X_{crop}$) is randomly cropped from a normal chest radiograph ($X_{normal}$) and randomly pasted at another place ($X_{paste}$). Simultaneously, an anatomical mask ($X_{mask}$) is created using element-wise multiplication between the segmented lung region ($X_{seg}$) and the blurred region ($X_{blur}$). Furthermore, $X_{normal}$ and $X_{paste}$ are summed by using $X_{mask}$. The anomalous region occurs only in the lung region in the augmented image.

## III. MATERIALS AND METHODS

### A. Overview

AnatPaste is an augmentation method that enables a model to learn the anatomical structural information. Fig. 2 depicts an AnatPaste augmentation method. Our proposed model first uses AnatPaste to train a convolutional neural network (CNN) model. Further, the trained model works as a feature extractor to detect anomalies. AnatPaste, the SSL training setup, and the anomaly score calculations are described below.

### B. AnatPaste augmentation

First, the following lung segmentation procedure is performed on the normal chest radiographs. CLAHE [46], [47] is used for image normalization, which improves visibility by homogenizing the image histograms and enhancing the local region contrast. The image is then binarized using Otsu's method [48]. Furthermore, noise is removed using morphological transformation [49]. The areas bordering the edges of the image are removed as background. The areas above a certain size are considered lung fields ($X_{seg}$ in Fig. 2).

Next, a patch ($X_{crop}$ in Fig. 2) is created by randomly cropping the original image $X_{normal}$. The size of $X_{crop}$ is the same as the original one, and the region other than the cropped region is valued at zero. The patch is randomly pasted at another place ($X_{paste}$). Then, an oval or rectangle is randomly drawn in the same position as the pasted position in $X_{paste}$ ($X_{blur}$). The figure's edge is blurred by a Gaussian blur to mimic the actual anomaly. This position is determined based on the following criteria: a) part of the figure should be included in the lung region, and b) the figure size should be smaller than the patch. An anatomical mask is created by calculating the element-wise multiplication between the segmented lung region and the blurred region ($X_{mask}$). Furthermore, $X_{normal}$ and $X_{paste}$ are summed by using an element-wise sum with $X_{mask}$. The pixel values of the blurred region and the Gaussian blurred radius are sampled randomly from uniform distributions of [0.59,1] and [0,15], respectively.

Given a normal image ($X_{normal}$), a patch ($X_{patch}$), and an anatomical mask ($X_{mask}$), the generated anomalous image ($X_{anomaly}$) is expressed as:

$$X_{anomaly} = X_{normal} * (1 - X_{mask}) + X_{patch} * X_{mask} \quad (1)$$

The loss function is expressed as:

$$\mathcal{L} = \frac{1}{N}\sum_{i=1}^{N} \{Lce(f(X_i), 0) + Lce(f(Aug(X_i)), 1)\} \quad (2)$$

wherein the ith $X_i$ of $N$ normal image data is input to $f$, a CNN model that makes binary predictions. Aug denotes AnatPaste augmentation. The original image is labeled 0, and the augmented one is labeled 1. Lce denotes the cross entropy loss [50]. Details of the lung segmentation is presented in Algorithm 1.

| | |
|---|---|
| **Algorithm 1:** Lung segmentation | |
| **Input:** | Normal chest radiograph |
| | Initialize $t$ of a threshold to be considered as lung region |
| **Output:** | Binarized lung segmentation image |
| step:1 | Equalize local image histogram using CLAHE. |
| step:2 | Binarize histogram equalized image using Otsu's thresholding. |
| step:3 | Apply morphological opening to remove noise. |
| step:4 | Remove objects connected to the image border. |
| step:5 | Apply morphological dilation to cover whole lung area: $X_{seg}$ |
| step:6 | Compute sets of data points $\{R_1 \dots R_K\}$ using connected-components labeling. Each set has same label. |
| step:7 | **for** k=1 : K<br>   **if** $|R_k| < t$ :<br>     $X_{seg} = g(X_{seg}, R_k, 0)$<br>   **end if**<br>**end for**<br>where $g(X, R_i, p)$ is a function that sets pixel value p for the pixels of the input image $X$ at coordinates in the $R_i$ and returns updated image. |
| step:8 | Return $X_{seg}$ |

*C. Anomaly Score*

We use kernel density estimator (KDE) [51] to calculate anomaly scores as reported by Li et al. [21]. Given the set of feature representations $D = \{z_i\}_{i=1}^{N}$ that the CNN outputs, KDE is computed as:

$$A(z) = \frac{1}{N}\sum_{i=1}^{N} K(z - z_i) \quad (3)$$

where

$$K(z) \propto \exp\left(-\frac{\|z\|^2}{2}\right) \quad (4)$$

The output anomaly scores are normalized from zero to one. A threshold is set to maximize the F1-score for each model.

IV. EXPERIMENTS

*A. Dataset*

Three publicly available chest radiograph datasets are used to verify performances. All the images are resized to 256 × 256 pixels.

*1) Zhanglab dataset [52]*

This dataset includes 6480 frontal view chest radiographs. Each image is labeled as normal or pneumonia (an anomalous condition). This dataset is officially split into training (1349 normal and 3883 pneumonia images) and test (234 normal and 390 pneumonia images) sets in advance. We choose 200 images (100 from normal and 100 from pneumonia) randomly from the training data for validating model selection and setting thresholds.

*2) Chexpert dataset [2]*

This dataset consists of 224,316 chest radiographs from 65,240 patients. The radiographs are annotated in 14 pathological categories, using radiology diagnostic reports. A total of 5,466 normal posterior-anterior (PA) images are used for training. We focus on Chexpert competition task diseases (Pleural Effusion, Cardiomegaly, Consolidation, Edema, and Atelectasis). A total of 300 normal and 300 abnormal images are selected randomly from the training dataset for the test. We select at least 50 images per disease. In the officially split validation dataset, 14 normal and 19 abnormal PA images are used for validation as suggested by Xiang et al. [18]. We ensure that patients do not overlap. We use this dataset in the ablation study. Some of the images are used in visualization..

*3) RSNA dataset [53]*

This is the Radiological Society of North America (RSNA) Pneumonia Detection Challenge dataset, which is a subset of the NIH chest radiograph dataset [54]. It consists of 9,790 normal and 19,894 abnormal (labeled as no lung opacity/not normal or lung opacity) chest radiographs. Of these, a total of 6,614 normal PA images are used for training, 300 normal and 300 abnormal images for validation, and 300 normal and 300 abnormal images for testing.

*B. Experimental Settings*

We use a one-class classifier with ResNet18 [55] for feature extraction. We add three fully connected layers to this backbone as the projection head. The batch size is set as 64. The network is optimized using an SGD optimizer [56], with an initial learning rate, momentum, and weight decay of 0.03, 0.9, and 0.00003, respectively. The learning rate is scheduled using a cycle cosine annealing schedule [57]. The model is trained for 256 epochs, and the weights are saved when the AUC for the validation dataset is the highest. The weights are loaded to evaluate performances on the test dataset. The entire framework is built on PyTorch [58] with an NVIDIA RTX 3090 24 GB GPU.

TABLE I
AVERAGE CLASSIFICATION PERFORMANCES (AUC, ACCURACY, F1) AND STANDARD DEVIATIONS YIELDED BY BASELINE MODELS AND OURS ON ZHANGLAB,
CHEXPERT, AND RSNA DATASET. ENSEMBLE MEANS THE AVERAGE ANOMALY SCORES OF FIVE ANATPASTE MODELS.

|  | Zhanglab | | | Chexpert | | | RSNA | | |
| --- | --- | --- | --- | --- | --- | --- | --- | --- | --- |
|  | AUC(%) | ACC(%) | F1(%) | AUC(%) | ACC(%) | F1(%) | AUC(%) | ACC(%) | F1(%) |
| fAnoGAN | 62.4±1.9 | 65.6±1.9 | 74.4±1.9 | 60.3±0.7 | 57.1±1.2 | 61.5±3.3 | 60.1±0.5 | 50.3±0.2 | 66.7±0.1 |
| Ganomaly | 75.1±1.2 | 66.8±5.7 | 78.5±3.0 | 61.7±4.6 | 56.5±4.0 | 46.5±20.6 | 65.7±9.8 | 52.7±2.1 | 67.4±0.8 |
| CutPaste | 76.8±2.9 | 75.2±2.4 | 81.4±1.6 | 58.0±1.3 | 52.8±2.0 | 64.5±1.6 | 69.1±2.0 | 51.6±3.0 | 66.7±0.1 |
| SALAD† | 82.7±0.8 | 75.9±0.9 | 82.1±0.3 |  |  |  |  |  |  |
| SQUID† | 87.6±1.5 | 80.3±1.3 | 84.7±0.8 |  |  |  |  |  |  |
| **AnatMix** | **91.4±1.3** | **83.0±1.9** | **86.8±1.1** | **76.0±1.8** | **65.4±3.8** | **71.2±1.4** | **80.4±1.4** | **72.3±1.8** | **75.1±0.7** |
| **Ensemble** | **92.1** | **83.2** | **86.4** | **78.7** | **69.2** | **73.5** | **81.9** | **72.8** | **75.6** |

† The results are taken from each paper.

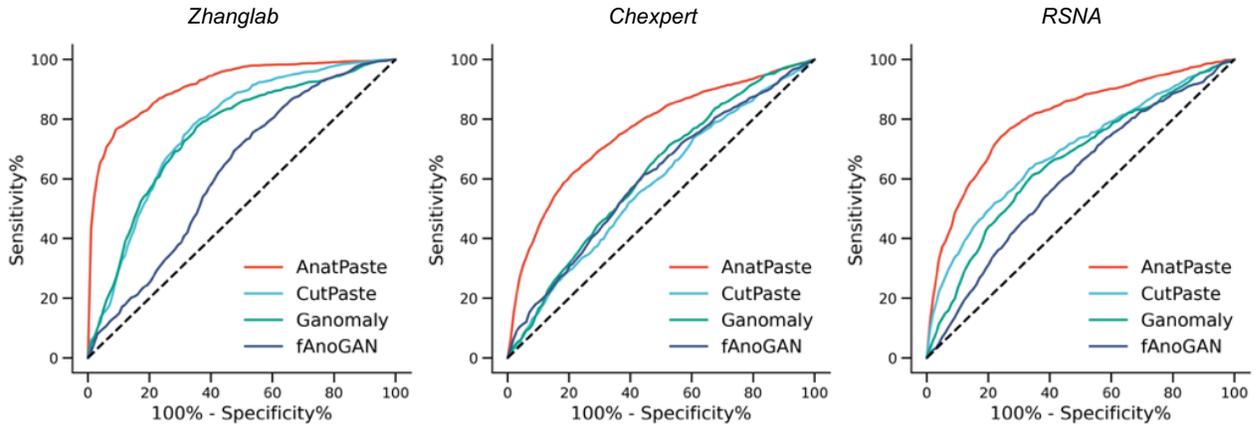

Fig. 3. Receiver operating characteristic curve (ROC) of the comparison methods on the three datasets

### C. Baseline & Evaluation

Several unsupervised anomaly detection models have been proposed. We compare the following three contemporary models.

- fAnoGAN [15] is the state-of-the-art UAD model in for medical images. Its three-component architecture contains an encoder, decoder, and discriminator. fAnoGAN is trained using only normal images. Abnormal images are detected when the decoder fails to restore them, leading to increased reconstruction errors.
- Ganomaly [38] is another GAN-based UAD model. It employs three losses (adversarial loss, contextual loss, and encoder loss) in anomaly detection. It has exhibited accurate anomaly detection in several datasets, including the X-ray security-screening datasets.
- CutPaste [21] is a one-class classifier model with SSL proposed for anomaly detection in industrial products. The model employs a pretext task to extract a region from a normal image, paste it on another part of the image, and classify the image as an anomaly. We use the

CutPaste-Scar variant, which exhibits the best accuracy in the original study.

We employ model-specific learning parameters as reported in the respective studies. For a fair comparison, all the methods except augmentation are identical in our model and the CutPaste algorithm. SALAD [19] and SQUID [18] are validated on the Zhanglab dataset. As the original code is unavailable, we compare them based on the accuracy reported in the respective papers. We validate the performances using the following metrics: receiver operating characteristic (ROC) curve, the area under curve (AUC), accuracy (ACC), and F1-score. We present the average scores out of five repeated runs. The validation threshold is determined based on the best F1-score.

## V. RESULTS

### A. Comparison With State of the art

Table I compares the contemporary methods' AUC, accuracy, and F1-scores on the three datasets. Our method exhibits better values than the contemporary methods. On the Zhanglab dataset, AnatPaste achieves 14.6% higher AUC than the conventional CutPaste (76.8%). The AUCs of SALAD and SQUID, which use the same test dataset for anomaly detection, are 82.7% and 87.6%, respectively. AnatPaste also outperforms these methods by more than 3.8% and achieves the best scores in accuracy and F1-score as well. Further, AnatPaste exhibits 18.0% and 11.3% higher AUC than the CutPaste algorithm on the Chexpert dataset and RSNA dataset, respectively, indicating that our method exhibits higher accuracy than the conventional models. Furthermore, the ensemble of the five AnatPaste models (Table I) achieves an accuracy improvement of 0.7%, 2.7%, and 1.5% on Zhanglab, Chexpert, and RSNA, respectively.

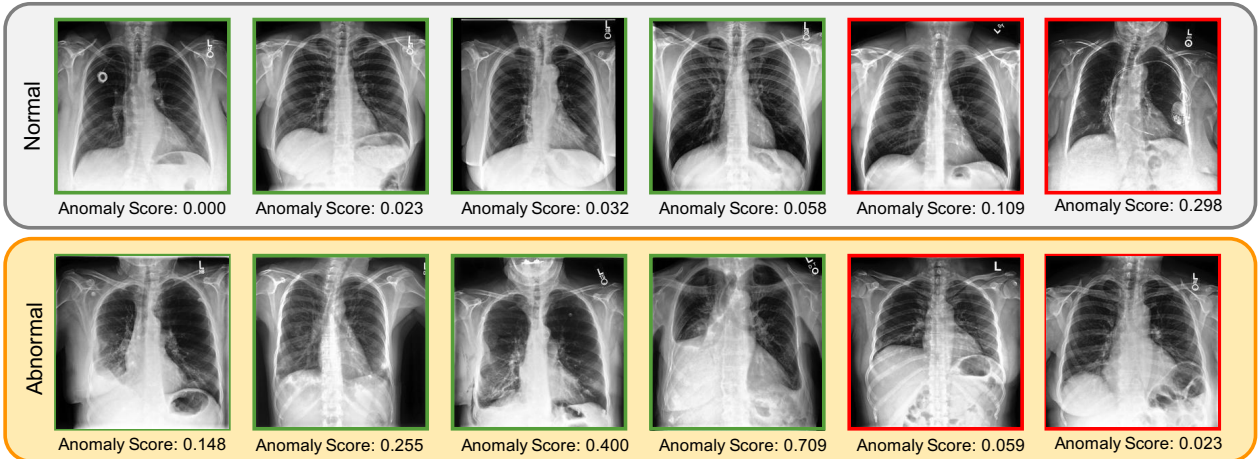

Fig. 4. Examples of the test dataset's predicted anomaly scores. The top and bottom rows denote normal and abnormal images, respectively. The anomaly scores are indicated below the corresponding images. Images marked with green boxes are recognized correctly by the model, while those marked with red boxes are not. The threshold is fixed to maximize corresponding F1-score.

### B. Anomaly score prediction

Anomaly scores are calculated for each image to assess their tendencies in normal and abnormal labels. The output of the model with the highest AUC score is shown in Fig. 4. The threshold is set to 0.067 based on the best F1-score. The images in green boxes are correctly predicted by the model, and those in red boxes are not. The more the areas of abnormal white shadows within the lung field, the higher the anomaly score tends to be. By contrast, the model often incorrectly predicts normal labeled images with pacemakers and other artifacts.

### C. Visualization

Fig. 5 depicts the Grad-CAM [59] visualization of the annotated areas in two images. The last convolutional layer's feature map of the pretext model is used to calculate the gradient. The original images (left panels) are enhanced using the Grad-CAM (right panels). The areas judged to be abnormal are highlighted in red. In the abnormal case, the red-colored highlighted areas appear corresponding to consolidation in the lung. By contrast, in the normal case, there are no red-colored annotated areas, and the model does not indicate a specific area.

### D. Evaluation by Disease

Table II shows the results of AnatPaste by disease on the Chexpert dataset. 300 images per disease and 300 normal images are randomly selected to validate performances. Pleural Effusion shows the best abnormality detection accuracy, followed by Consolidation and Edema. The AUC of consolidation, edema, and pleural effusion achieved via supervised learning in the original paper is similar by approximately 93% [2]. Compared to these results, AnatPaste has a relatively high accuracy in detecting Pleural Effusion.

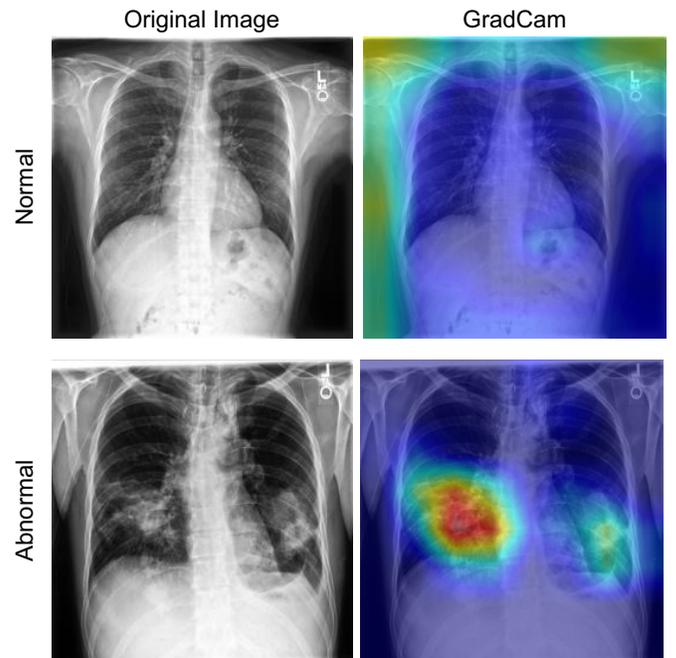

Fig. 5. Grad-CAM visualizations of normal and abnormal chest radiograph images. The first and second columns depict the original images and Grad-CAM visualizations, respectively. Red areas in Grad-CAM visualizations highly contribute to the prediction.

TABLE II
ANOMALY DETECTION PERFORMANCE BY DISEASE IN CHEXPERT DATASET.

|  | AUC(%) | ACC(%) | F1(%) |
| --- | --- | --- | --- |
| Atelectasis | 71.2±1.8 | 63.1±3.2 | 68.7±0.5 |
| Cardiomegaly | 73.3±1.8 | 62.9±3.9 | 68.4±1.6 |
| Consolidation | 78.2±2.1 | 65.6±4.1 | 71.5±2.0 |
| Edema | 76.5±1.7 | 64.7±4.3 | 70.5±2.0 |
| Pleural Effusion | 80.0±2.4 | 66.8±4.4 | 72.8±2.0 |

### E. Ablation Study

We perform an ablation study to validate the essentiality of the components of AnatPaste, which are lung segmentation and Gaussian blur. We investigate the changes in accuracy after their removal. The Chexpert dataset is used for the validation. The evaluation metrics (AUC, accuracy, and F1-score) are calculated. In Table III, the AUCs without segmentation and without Gaussian blur are 70.7% and 60.2%, respectively, which are 5.3% and 15.8% lower than that of AnatPaste, respectively, suggesting that both the components improved accuracy.

TABLE III.
ABLATION STUDY COMPARING THE FULL ANATPASTE, ANATPASTE WITHOUT IMAGE SEGMENTATION, AND ANATPASTE WITHOUT GAUSSIAN BLUR.

|  | AUC(%) | ACC(%) | F1(%) |
|---|---|---|---|
| Full AnatPaste | 76.0±1.8 | 65.4±3.8 | 71.2±1.4 |
| w/o Segmentation | 70.7±1.6 | 55.7±5.9 | 66.2±0.9 |
| w/o Gaussian blur | 60.2±2.9 | 52.5±2.7 | 65.1±2.5 |

## VI. DISCUSSION

Chest radiography is the most common and minimally invasive screening test in medical diagnosis. There are various chest diseases to be detected, but the shortage of labeled datasets has prevented accurate training of disease detection models. The UAD model not only solves such problems but also serves as a valuable computer aided diagnosis system that provides appropriate treatment decisions.

In this study, we propose a simple and effective SSL-based method that detects chest radiograph anomalies, using anatomical structures in the lung fields. Our method takes advantage of the medical knowledge that most abnormalities appear in the lung field region and augments only the segmented lung fields. The images generated by our method are more similar to the actual anomalies than those generated by existing methods. Our model trained with AnatPaste exhibits high feature representation capability even without disease labels and significantly improves anomaly detection accuracy at least by 3.8%, 14.3%, and 11.3% in terms of AUC compared with existing methods on three publicly available datasets.

To the best of our knowledge, our model is the first SSL-based model that employs anatomy-aware segmentation tasks. Our method employs the same network architecture, loss function, and training scheme as previous methods, which suggests that AnatPaste improves anomaly detection accuracy. Our results indicate that incorporating disease and anomaly-specific medical expertise in the training can enable accurate medical image analysis.

AnatPaste uses segmentation methods based on traditional methods such as Otsu's binarization to extract anatomical structures. Since AnatPaste is applied only to normal images, we think that it should sufficiently extract lung regions. In addition, other deep learning-based lung segmentation methods require annotations and are outside the scope of the unsupervised learning framework. AnatPaste can be extended to other organs and modalities by employing appropriate segmentation methods.

AnatPaste can be applied as a pretext task regardless of the structure of the deep learning model; therefore, further performance improvement can be expected by incorporating it into existing anomaly detection models. In addition, AnatPaste's feature representation capability can be used not only for anomaly detection but also for downstream tasks such as disease classification and segmentation through supervised learning. Thus, AnatPaste can be applied to a variety of medical image analysis and has the potential to contribute to improving accuracy.

## VII. CONCLUSION

We proposed a novel SSL-based method for chest radiograph anomaly detection. Our objective was to adopt carefully designed augmentations such that abnormal areas are limited to a single organ, thus allowing the model to recognize anomalies in the context of the anatomical structure. Experimental results on three chest radiograph benchmark datasets demonstrated that our method outperformed existing unsupervised anomaly detection methods. AnatPaste has the potential to improve the performance of any deep learning model, in any medical image modalities, and for any organs.